\documentclass[default,iicol]{sn-jnl}

\usepackage{graphicx}%
\usepackage{multirow}%
\usepackage{amsmath,amssymb,amsfonts}%
\usepackage{amsthm}%
\usepackage{mathrsfs}%
\usepackage[title]{appendix}%
\usepackage{xcolor}%
\usepackage{textcomp}%
\usepackage{manyfoot}%
\usepackage{booktabs}%
\usepackage{algorithm}%
\usepackage{algorithmicx}%
\usepackage{algpseudocode}%
\usepackage{listings}%
\usepackage{geometry}

\usepackage{bbding}
\usepackage{times}
\usepackage{epsfig}

\theoremstyle{thmstyleone}%
\theoremstyle{thmstyletwo}%

\theoremstyle{thmstylethree}%

\begin{document}

\title[Article Title]{A Simple Framework for 3D Occupancy Estimation in Autonomous Driving}


\author[1,2]{\fnm{Wanshui} \sur{GAN}}\email{wanshuigan@gmail.com}

\author[3]{\fnm{Ningkai} \sur{Mo}}\email{nk.mo19941001@gmail.com}

\author[4]{\fnm{Hongbin} \sur{Xu}}\email{hongbinxu1013@gmail.com}

\author[1,2]{\fnm{Naoto} \sur{Yokoya}}\email{yokoya@k.u-tokyo.ac.jp}
\equalcont{Corresponding author.}

\affil[1]{\orgname{The University of Tokyo}, \orgaddress{ \state{Tokyo}, \country{Japan}}}

\affil[2]{\orgname{RIKEN }, \orgaddress{\state{Tokyo}, \country{Japan}}}

\affil[3]{\orgname{Shenzhen Institute of Advanced Technology, Chinese Academy of Sciences}, \orgaddress{ \state{Shenzhen}, \country{China}}}

\affil[4]{\orgname{South China University of Technology}, \orgaddress{\state{Guangzhou}, \country{China}}}


\abstract{The task of estimating 3D occupancy from surrounding-view images is an exciting development in the field of autonomous driving, following the success of Bird's Eye View (BEV) perception. This task provides crucial 3D attributes of the driving environment, enhancing the overall understanding and perception of the surrounding space. In this work, we present a simple framework for 3D occupancy estimation, which is a CNN-based framework designed to reveal several key factors for 3D occupancy estimation, such as network design, optimization, and evaluation. In addition, we explore the relationship between 3D occupancy estimation and other related tasks, such as monocular depth estimation and 3D reconstruction, which could advance the study of 3D perception in autonomous driving. For evaluation, we propose a simple sampling strategy to define the metric for occupancy evaluation, which is flexible for current public datasets. Moreover, we establish a benchmark in terms of the depth estimation metric, where we compare our proposed method with monocular depth estimation methods on the DDAD and Nuscenes datasets and achieves competitive performance. The relevant code is available in \href{https://github.com/GANWANSHUI/SimpleOccupancy}{https://github.com/GANWANSHUI/SimpleOccupancy}.}




\maketitle

\section{Introduction}
\label{sec:intro}

3D scene understanding is a challenging mission for autonomous driving, especially only relying on the camera. In recent years, Bird's Eye View (BEV) perception, including 3D detection \cite{liu2022petr} and map segmentation \cite{li2022hdmapnet} is getting a lot of attention with the advantage of doing the 3D task estimation in the 2D feature plane and is beneficial for downstream tasks such as prediction and planning \cite{bevsurvey, arnold2019survey}. However, some vital information for driving safety is ignored in the BEV tasks, such as an unrecognizable obstacle. Therefore, reconstructing the 3D geometry of the driving scenes is a longstanding task for autonomous driving. 

To obtain the 3D geometry information, depth estimation from RGB images, such as monocular depth estimation \cite{monosurvery, xiang2023towards,lee2022self} and stereo matching \cite{poggi2021synergies}, has been well investigated. While depth maps can provide 3D geometry information at the pixel level, we need to project them into the point cloud format in 3D space, and multiple post-processing procedures are required as depth maps may be inconsistent in the local region \cite{tesla}, which is not a straightforward manner for the 3D perception in autonomous driving. For better geometry representation in driving scenarios, occupancy estimation has gained attention in the industrial community. It has shown superiority over the representation in the BEV space \cite{tesla}. To advance the study in 3D occupancy estimation, we explore a simple framework for 3D occupancy estimation, starting from the surrounding-view setting, like BEV perception. 

\begin{figure}[t!]
\centering
{
\includegraphics[width=\linewidth]{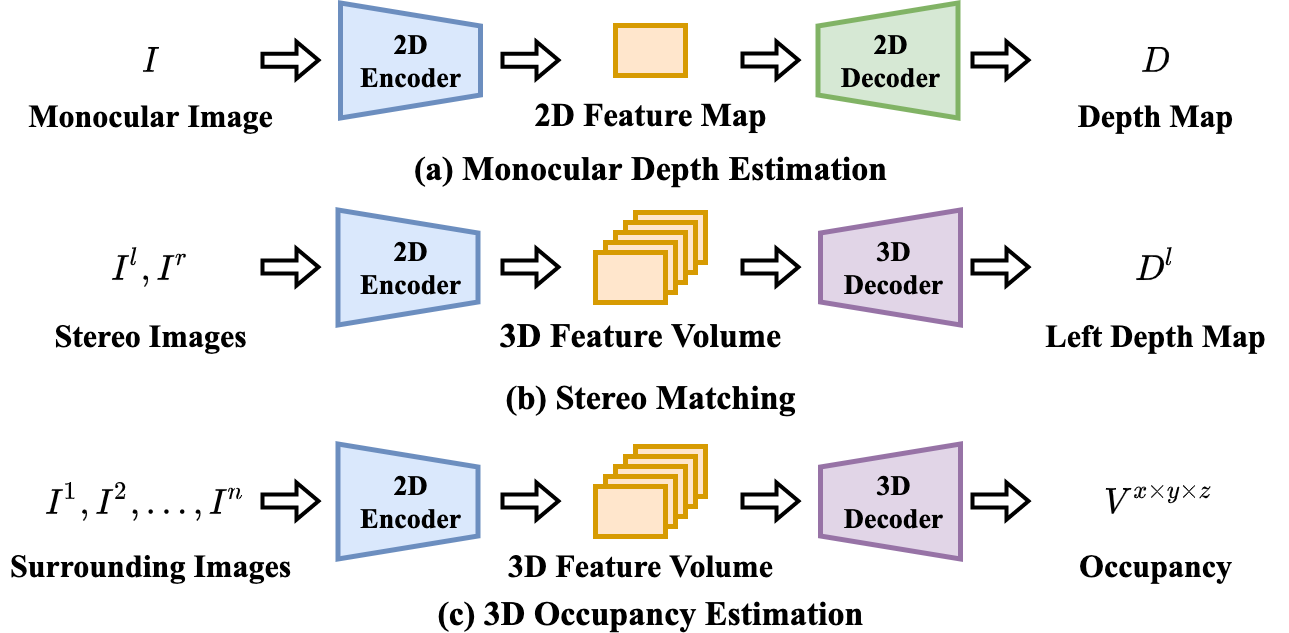}
}
\caption{\textbf{A comparison of the overall pipline of monocular depth estimation, stereo matching, and 3D occupancy estimation}.}
\label{fig:task}
\end{figure}

To make the framework self-contained, we investigate the baseline in terms of network design, optimization, and evaluation. For the network design, as shown in Figure \ref{fig:task}, the final output representation for 3D occupancy estimation is different from monocular depth estimation and stereo matching. The network architecture of the 3D occupancy estimation is similar to stereo matching, which means that the experiences from the stereo matching task can be adapted to occupancy estimation to reduce the burden in the network design. Therefore, we design the pipeline to resemble stereo matching closely and investigate a CNN-based framework as the baseline. In terms of optimization, we investigate supervised and self-supervised learning, which are based on the render depth map \cite{nerf} and point-level classification label. For evaluation, we conduct experiments on two well-known datasets, DDAD and Nuscenes \cite{ddad, nuscenes}, for broader recognition. Besides, we propose a novel distance-based metric for occupancy evaluation, which is inspired by the sampling strategy in volume rendering \cite{nerf}. Our experimental results demonstrate that the proposed metric is more equitable compared to alternative options, such as classification metrics. Additionally, the proposed metric boasts flexibility as it solely relies on the point cloud as the ground truth, thereby eliminating any additional burdens when implementing the metric on similar datasets.

The main contributions of this work are summarized as follows.
\begin{itemize}
    \setlength{\itemsep}{0pt}
    \setlength{\parsep}{0pt}
    \setlength{\parskip}{0pt}
    \setlength{\topsep}{0pt}
    \setlength{\partopsep}{0pt}
    \item We introduce a framework for surrounding-view 3D occupancy estimation, featuring a novel network design, loss design, and evaluation metric based on discrete point level sampling.
    \item We propose an occupancy metric and demonstrate its efficacy for 3D occupancy evaluation. Furthermore, we connect 3D occupancy estimation to the monocular depth estimation task and establish a new depth estimation benchmark with competitive performance.
    \item We delve deeper into self-supervised learning within our framework, extending its application to 3D reconstruction through the representation of the signed distance function. To the best of our knowledge, we are the first work to investigate the 3D reconstruction at mesh level in the surrounding-view driving scenes.
    \item We propose an effective pretrain strategy for the 3D semantic occupancy task based on the sampling strategy and reveal the different characteristic of the point-level optimization and dense voxel-level optimization.
    \item Through extensive qualitative and quantitative experiments for the DDAD and Nuscenes datasets, we demonstrate the effectiveness of our proposed framework as a universal solution that can utilize existing unlabeled point cloud datasets to perform the 3D occupancy estimation and evaluation as simple as depth estimation. 
    
\end{itemize}

\section{Related Work}

\subsection{Depth estimation}
For monocular depth estimation, it is usually implemented with a 2D U-Net architecture, as Figure \ref{fig:task} shown \cite{monodepth2,newcrfs}. More recently, the surrounding-view depth estimation has been explored, which is not just limited to a single image context \cite{fsm,surrounddepth}. FSM \cite{fsm} uses the spatio-temporal contexts, poses consistency constraints, and designed photometric loss to learn a single network generating dense and scale-aware depth map. Differently, Surrounddepth \cite{surrounddepth} adopts structure-from-motion to extract scale-aware pseudo depths to pretrain the models for the scale-aware result. This work mainly discusses 3D occupancy estimation, but we compare the depth metric with monocular depth estimation methods. 

In terms of stereo matching, we could obtain the real scale depth map by estimating disparity between the stereo images. The state-of-the-art methods usually use 3D convolution neural networks (CNN) to do cost aggregation \cite{leastereo,cspn,xu2022attention,psmnet,HybirdNet}. Likewise, we also use the 3D CNN to do 3D volume aggregation for the final occupancy representation. 


\subsection{BEV perception}
We identify that the perception task from the bird's eye view has a common step as the 3D occupancy estimation. Both of them require feature space transformation, where the BEV perception task is from the image space to the BEV plane \cite{bevsurvey}, and 3D occupancy estimation is from the image space to the 3D volume space. LSS \cite{lss} implicitly projects the image feature to the 3D space by learning the latent depth distribution. DETR3D \cite{detr3d} and BEVFormer \cite{bevformer} define a 3D query in the 3D space and use the transformer attention mechanism to query the 2D image feature. ImVoxelNet \cite{imvoxelnet} and Simple-BEV \cite{simplebev} build the 3D volume by the bilinear interpolation of the projection location at the 2D image feature plane, which is an efficient and parameter-free unprojection manner. Therefore, for simplicity in this baseline work, we adapt the parameter-free unprojection manner to build the 3D volume the same as Simple-BEV. 

\begin{figure*}
    \centering
    \includegraphics[width=0.99 \textwidth]{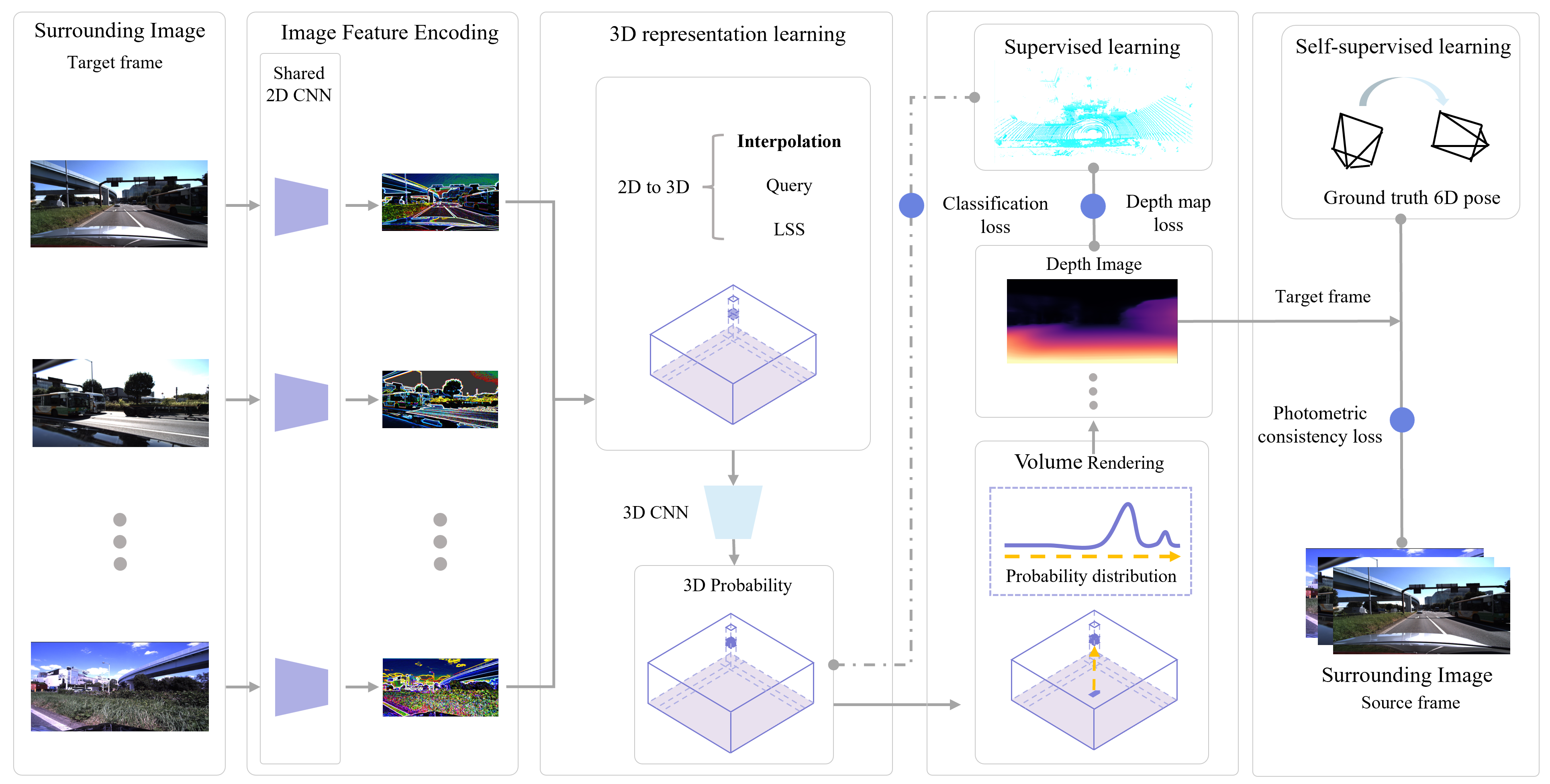}
    \vspace{0.2cm}
    \caption{\textbf{The overview of the proposed Simple 3D Occupancy estimation framework (SimpleOccupancy).} Given the surrounding image, we first extract the image feature by the shared 2D CNN and then use the parameter-free interpolation to obtain the 3D volume. The following 3D CNN could effectively aggregate the 3D feature in the volume space (Section \ref{3.2}). At last, we train the proposed network for both the supervised learning from the sparse point cloud and the self-supervised learning with photometric consistency loss. (Section \ref{3.4}).}
    \label{fig:Overview}
\end{figure*}

\subsection{Image-based 3D reconstruction}
There have been a series of methods for rendering-based 3D reconstruction \cite{scenerf, behindthescenes,volrecon,sat2density, streetsurf}. VolRecon \cite{volrecon} and StreetSurf \cite{streetsurf} use the signed distance function for the scene reconstruction, while VolRecon requires the multi-view image as the input and StreetSurf is not a generalizing method, only for the single scene reconstruction. Sat2density \cite{sat2density} uses volume rendering to transform the satellite image into the density to help the ground view synthesis. The recent works \cite{scenerf, behindthescenes} explore the volume rendering for the depth estimation in single-view image in the autonomous driving scene. Different from them, our task is more challenging for the 3D reconstruction from the surrounding-view image and we investigate the proposed framework in a more comprehensive manner, ranging from supervised and self-supervised learning, and also, jointly investigating the 3D occupancy estimation, depth estimation, and 3D reconstruction.

\subsection{Occupancy estimation}
There have been some works representing the scene in an occupancy (voxel) format \cite{occupancysurvey,mescheder2019occupancy,lange2022lopr, huang2023tri}. Recently, the industrial community \cite{tesla} reveals that the occupancy representation could be easily combined with the semantic information achieving instance-level prediction. Besides, it could predict the occupancy flow by considering the sequence frame, which could be regarded as the scene flow in the instance level \cite{sceneflow}. Following \cite{tesla}, this personal blog gives a try to use volume rendering to train the occupancy representation in a self-supervision manner, which can not reconstruct the dynamic object and requires the velocity information of the vehicle \cite{VoxelNet}. For the geometry of driving scene, MonoScene \cite{monoscene} and Voxformer \cite{voxformer} explore the 3D semantic scene completion from a single image. This approach proposes to reconstruct the road scene from a monocular image by the deep implicit function \cite{roddick2021road}. Different from \cite{monoscene,roddick2021road, voxformer}, we are under the surrounding-view setting, which is more challenging and meets the need for full perception in the driving scenarios. We need to point out that this baseline work is inspired by the above two works \cite{tesla,VoxelNet} from industrial community, where we contribute a new CNN-based framework, a novel evaluation metric and set up the benchmark for both the 3D occupancy metric and depth map metric in the public datasets \cite{ddad, nuscenes}. 
    
Upon completing this study, we discovered several relevant preprint works \cite{wei2023surroundocc,wang2023openoccupancy, tong2023scene}, exploring the 3D occupancy estimation in a surrounding-view setting. These works share a similar motivation of constructing dense voxel labels for semantic occupancy prediction, which involves merging multi-frame point clouds with human annotations to handle dynamic objects and extensive post-processing. In contrast, our focus is solely on pure geometry prediction without incorporating semantic information. Our method operates at a point level based on the sampling technique depicted in Figure \ref{fig:label}. This difference allows our work to do the 3D reconstruction at the mesh level and offers pretraining benefits to enhance the concurrent works, as presented in the experiment section.

\section{Method}

\subsection{Preliminaries}

In this paper, we adopt volume rendering to obtain the depth map for the model training. In the novel view synthesis task \cite{nerf, v4d}, researchers usually use a multilayer perceptron network (MLP) to learn a mapping function to map the 3D position $(x,y,z)$ and view direction $(\theta, \phi)$ into the volume density $\sigma$ and the RGB color $c$. For rendering the RGB value in a pixel, we can use the approximate volume rendering in NeRF \cite{nerf} to do the weighted sum on the sampled point on the ray. The rendering function is defined in the equation (\ref{rendering}): 
\begin{equation}
    \hat{c} = \sum_{i=1}^{W}{T_{i}(1-exp({-\sigma _{i}}{\delta _{i}}  )  ) c_{i} }, 
     \label{rendering}
\end{equation} 
where $\hat{c}$ is the rendered RGB value, ${T_{i} = exp\left ( -\sum_{j=1}^{i-1} {\sigma _{j}}{\delta _{j}}  \right ) }, {\delta _{i}} = t_{i+1} - t_{i}$ is the adjacent sampled points' distance, $i$ is the sampled point along the ray, and $W$ is the number of the sampled points. If we want to obtain the depth information of the related pixel, we can replace the RGB value in the equation (\ref{rendering}) with the sampled point's distance $t_{i}$ as shown in equation (\ref{weight}). In this way, we can use the ground truth depth map to do the supervision training. 
 \begin{equation}
    \hat{d} = \sum_{i=1}^{W}{T_{i}(1-exp({-\sigma _{i}}{\delta _{i}}  ) )t_{i} }.
     \label{weight}
\end{equation} 
The NeRF's model learns geometry by the multi-view constraint while, in our setting, the geometry is from the image that is achieved by the CNN model, as introduced below.

\subsection{Model design} \label{3.2}

The problem setting of this work is defined in Figure \ref{fig:task}. Given the surrounding-view images with the relative camera pose to the vehicle framework, we design an end-to-end neural network $Q$ to predict the 3D occupancy map, where we formulate it as  $Q$: $(I^1,I^2,I^3,..., I^n) \rightarrow V^{x \times y \times z}$, where $n$ is the number of the surrounding images and $x,y,z$ represents the resolution of the final output of the voxel. We present the overview of the proposed framework in Figure \ref{fig:Overview}, and give the details as follows.


\textbf{Encoder.} For the image feature extraction, we use the ResNet \cite{resnet} as the backbone. We provide the experiment on ResNet 50 and 101 with the pre-trained model from ImageNet \cite{pytorch, krizhevsky2012imagenet}. The final feature map is with the shape $C \times H/4 \times W/4 $, where $H$ and $W$ are the input resolution of the image and $C=64$ is the channel number. 

\textbf{From the image feature to 3D volume.} For both the BEV perception and 3D occupancy estimation, a critical step is to transform the image feature from the 2D image space to the 3D volume space. In this work, we adopt the simplest manner as used in the Simple-BEV \cite{simplebev}. Specifically, we define a set of 3D points and project the 3D points to the 2D image feature planes and do the sampling by the bilinear interpolation. For the overlap region of the adjacent camera, we do the mean average for the sampled feature. Besides, we also investigate the other two 2D to 3D transformations, LSS \cite{lss} and Query \cite{wei2023surroundocc}, but we do not observe the benefit of using them as shown in the experiment section.

\textbf{3D volume space learning.} The parameter-free transformation by the bilinear interpolation does not have any position prior, which means that the feature on the rays of the frustum is identical. Therefore, it is a highly ill-posed setting to infer the 3D geometry from the surrounding-view images with only a little overlap. The ill-posed setting requires stronger feature aggregation ability to achieve the 3D occupancy prediction. For the 3D feature learning, we adapt the 3D convolution network based on the hourglass structure from HybridNet \cite{HybirdNet} in the stereo-matching task. Differently, we discard the multiple 3D volume output from the stacked hourglass architecture because we do not find the performance improvement with the default architecture, and rendering for the multiple depth maps also needs more computational resources. The detailed network structure is placed in the supplementary material.  

\textbf{Occupancy probability.} After the 3D volume space feature aggregation, we have the final voxel $V^{x \times y \times z}$. The occupancy probability value is from the Sigmoid function (\ref{sigomid}). 
\begin{equation}
    Probability = Sigmoid(\sigma).
     \label{sigomid}
\end{equation}

\textbf{Signed distance function (SDF).} Building upon prior research \cite{wang2021neus, Oechsle2021ICCV}, it has become evident that relying on density (probability) derived from volume rendering presents challenges as a reliable geometric representation, owing to its intrinsic geometric inaccuracies, notably biases in surface reconstruction. To address this issue, we delve deeper into the exploration of rendering-based 3D reconstruction, specifically employing the signed distance function (SDF) within a surrounding-view context. Following the methodology outlined in Monosdf \cite{yu2022monosdf} and VolSDF \cite{yariv2021volume}, we adopt a parallel approach to convert SDF values $s$ into density $\sigma$ before the volume rendering process as follows.
\begin{equation}
\sigma_\beta(s)=\left\{\begin{array}{ll}\frac{1}{2 \beta} \exp \left(\frac{s}{\beta}\right) & \text { if } s \leq 0 \\ \frac{1}{\beta}\left(1-\frac{1}{2} \exp \left(-\frac{s}{\beta}\right)\right) & \text { if } s>0\end{array}\right.,
\end{equation} 
where $\beta$ is the learnable parameter, initialized with 1.0.

\subsection{Model evaluation} \label{3.3}

\begin{figure}[t!]
\centering
{
\includegraphics[width=0.48\textwidth]{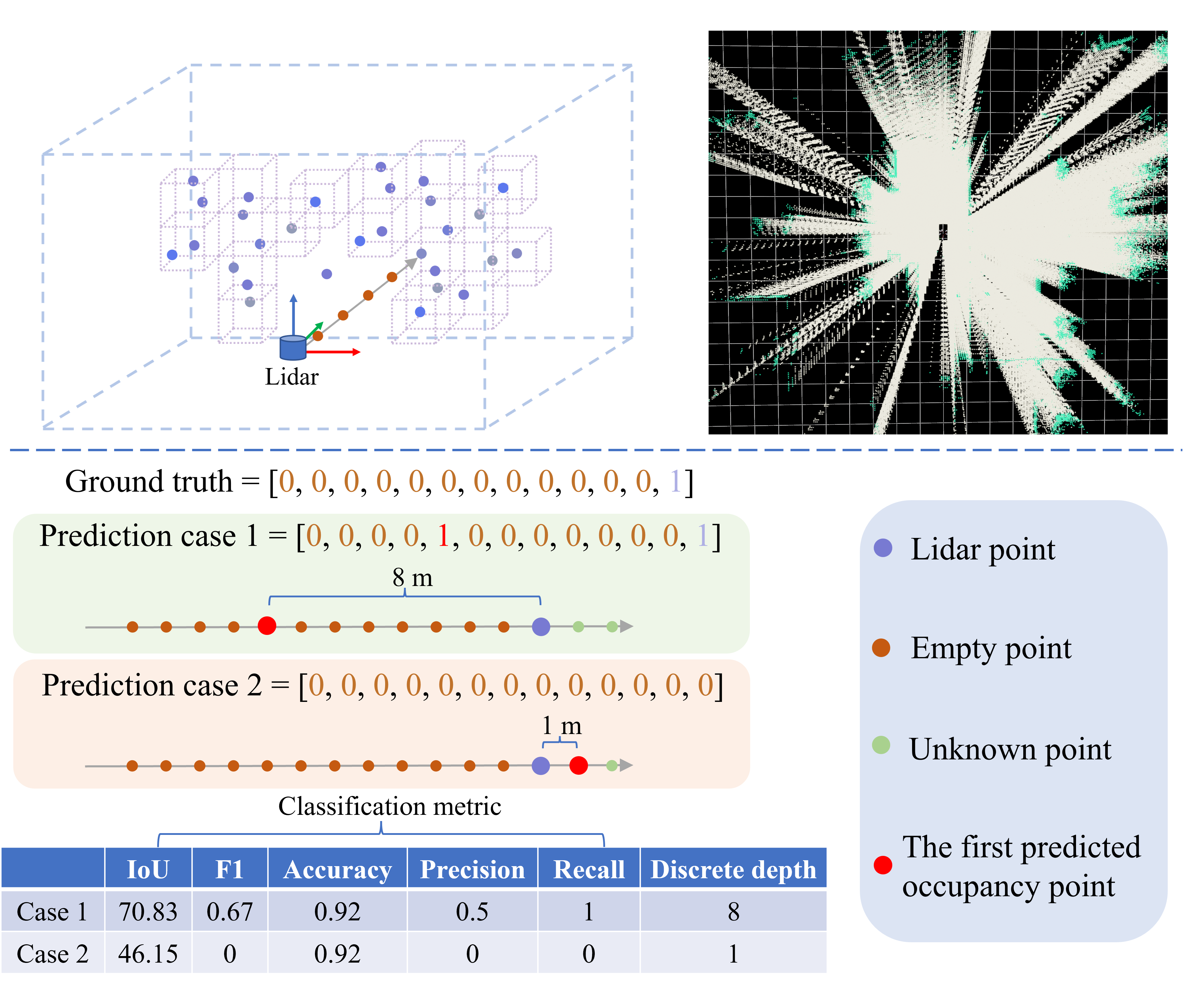}
}
\caption{\textbf{We use a collection of key points to represent the 3D space and evaluate it based on sample points.} The classification label is shown on the upper side, while the bottom side compares two metrics - the classification metric and our newly proposed discrete depth metric - in two different prediction cases. Note that the unknown point is not involved in classification metric but in the proposed discrete depth metric. It is evident that the discrete depth metric accurately reflects the cost associated with each prediction.}
\label{fig:label}
\end{figure}

To the best of our knowledge, our study represents pioneering research in the field of surrounding-view 3D occupancy estimation and concurrent evaluation. As a result, we find it imperative to explore suitable metrics for assessing the performance of our model, particularly considering the available datasets - Nuscenes and DDAD datasets. Our approach aims to leverage existing datasets for 3D occupancy estimation and evaluation in a manner similar to depth estimation - with simplicity and efficiency in mind. To achieve this goal, we have identified key factors and essential considerations that should be taken into account during the evaluation process:
(1) The current available outdoor dataset with the point cloud as the ground truth label is sparse, especially for the far-end space. Without huge efforts of human labeling, we can not obtain the dense voxel label as in \cite{semantickitti}. Therefore, we can not evaluate the whole voxel space and could only do the evaluation on the known space. (2) About the known space, we can only determine the space between the lidar center and the point cloud. (3) The 3D occupancy and voxel representation is a discrete representation, which means the quantization error is unavoidable, but we can determine the affordable quantization error in the autonomous driving scenario. (4) The evaluation should be feasible for the common datasets and easily be conducted for the study. Therefore, given the aforementioned considerations, we examine two evaluation metrics in this study: the classification metric and the discrete depth metric. These metrics are also associated with two supervised training fashions in the ablation study as shown in Figure \ref{fig:label}.

\textbf{Occupancy label generation.} Inspired by NeRF \cite{nerf}, we use the stratified sampling strategy to set the label for the free space. First, we set Lidar's position as the origin of all the rays for all the point clouds. Different from the even stride sampling, we use a fixed number of sample points (30), which means that, for closer point clouds, the sampling space is denser, and for far-end point clouds, the sampling space is sparser. The used sampling strategy has advantages in that it allows for a more balanced distribution of positive and negative labels and maintains a higher standard for closer objects in the driving scenes. The occupied space is represented by point cloud voxelization with a resolution of 0.4 m. The above operation can be easily performed using the Open3D \cite{open3d} library. 

\textbf{Classification metric.} By representing the known space by a set of key points, we can perform the evaluation as the classification task with binary classification metrics, as shown in Figure \ref{fig:label}. The classification metrics are commonly used in 3D semantic scene completion tasks \cite{occupancysurvey, monoscene, voxformer}. However, we observe that classification metrics are not perfect since they can only evaluate the known space in our setting. We give two cases of evaluation along a ray in Figure \ref{fig:label}. We can see that classification metric is not sensitive to case 1. Even though the first predicted occupancy point is far away from the ground truth occupancy (lidar point), classification metric still gives a high score. Conversely, in case 2, if the network is unable to predict the first occupancy point in known space, all of the classification metrics produce a low score except for the accuracy metric. This is unfair if the first predicted occupancy point is next to the actual occupancy point. For a more detailed explanation, please refer to the supplementary material.

\textbf{Discrete depth metric.} Recognizing the limitations of classification metric, we introduce discrete depth metric, which provides a more accurate assessment of predictions. Our approach involves dense sampling evaluation points along the ray with an interval (0.2 m) and a maximum distance (52 m). If all the prediction along the ray is empty, we set the last point as the first predicted occupancy point.
The discrete depth error is then calculated as the distance between the first predicted occupancy point and point cloud along the ray. By utilizing this criterion, we can perform evaluations in a manner similar to depth map error assessments. Following depth estimation \cite{surrounddepth}, we report occupancy evaluation with the following metrics, including error metric, (Abs Rel, Sq Rel, RMSE, RMSE log) and accuracy metric, $\delta<t: \% \text { of } d \text { s.t. } \max \left(\frac{\hat{d}}{d^*}, \frac{d^*}{\hat{d}}\right)=\delta<t$. The detailed definition is presented in supplementary material.

\subsection{Model optimization} \label{3.4}
In this work, we mainly utilize the depth map from volume rendering for the training, which involves supervision learning with ground truth labels (depth maps and the generated point labels) and self-supervised learning with the photometric reconstruction loss \cite{monodepth2}.
\subsubsection{Supervised learning}
Based on the available depth map and the generated occupancy label as introduced before, we investigate two different training manners. The first one is depth loss, the depth map supervision by the volume rendering. The other one directly calculates the binary classification loss based on the obtained label in the known space, where we consider binary cross entropy loss and L1 loss.

\textbf{Depth map loss.} With the depth map from the volume rendering, we can train the network the same as the depth estimation task. Following \cite{newcrfs}, we use the Scale-Invariant Logarithmic (SILog) loss \cite{eigen2014depth} to supervise the training. Specifically, the depth map loss is defined as: 
\begin{equation}
\mathcal{L}_{depth}=\alpha \sqrt{\frac{1}{M} \sum_i \Delta d_i^2-\frac{\lambda}{M^2}\left(\sum_i \Delta d_i\right)^2},
\end{equation}
where $\Delta d_i=\log \hat{d}_i-\log d_i^*$, $d_i^*$ is the ground truth depth and $\hat{d}_i$ is the predicted depth. $M$ is the number of valid pixels. We set $\lambda = 0.85$ and $\alpha = 10$ the same as \cite{newcrfs}.

\textbf{Classification loss.} Following the common practice, we use the binary cross entropy (BCE) loss function to train the model as follows: 
\begin{equation}
\mathcal{L}_{BCE}=-\frac{1}{N} \sum_{i=1}^N y_i \cdot \log \left(\hat{y_i}\right)+\left(1-y_i\right) \cdot \log \left(1-\hat{y_i}\right),
\end{equation}
where $y_i$ is the ground truth label and $\hat{y_i}$ is the binary probability prediction. Note that we used to try the combination of the BCE loss and Dice loss \cite{dice} to balance the occupied and empty labels, but we did not observe the improvement. Besides, we also investigate to use of the L1 loss direct work on sampled points as below: 
\begin{equation}
\mathcal{L}_{L1}=\frac{1}{N}\sum_{i=1}^{N}L_{1}\left ( 1-p_{i}  \right ) + \frac{1}{K} \omega \sum_{j=1}^{K}L_{1}\left ( 0-p_{j}  \right ),        
\end{equation}
where $p_{i}$ is the probability value based on the point cloud position, and $p_{j}$ is the probability value from the sampled point in the empty space. $N$ is the number of the valid point cloud and $K$ is the number of sampled points in the empty space. $\omega = 5 $ is the hyperparameter for balancing occupied and empty labels with the search range from 1.0 to 10.0.

\subsubsection{Self-supervised learning}
In the self-supervised learning setting, we use the photometric consistency loss for the training, which follows the \cite{monodepth2, surrounddepth}. We formulate the loss function in: 
\begin{equation}
\label{self-supervised loss}
\mathcal{L}_{self}\left(I_t, \hat{I}_t\right)=\beta \frac{1-\operatorname{SSIM}\left(I_t, \hat{I}_t\right)}{2}+(1-\beta)\left\|I_t-\hat{I}_t\right\|,
\end{equation}
where the $I_t$ and $\hat{I}_t$ mean the target image and the corresponding synthesized image. $\hat{I}_t$ is generated by the pixels projected from the source image to the target images $I_t$. $\beta$ is set as 0.85 for balancing two loss terms. Note that the projection requires the camera intrinsic matrix $K$ and the relative transformation pose for $T$ the same as the in \cite{surrounddepth, monodepth2}. Here, we use the ground truth pose collected by the sensor rather than the pose network to predict the pose. Because in our rendering setting, we predict the real-world scale that is difficult to learn from the pose network shown in the \cite{surrounddepth}. We begin the study of the rendering-based surrounding-view self-supervised learning with the ground truth pose, considering the sensor relatively easily obtains it and also has been widely used in the temporal module study \cite{bevformer, liu2022petrv2}.

\begin{table*}[t]
\caption{The ablation study of the proposed method in the proposed \textbf{discrete depth metric} for occupancy evaluation. Experiment (7) means do not use the pretrained model from ImageNet. Experiment (6) is the optimal setting, and we regard it as the full model. $\uparrow$ means the value higher is better and $\downarrow$  means lower is better. The number with bold typeface means the best and the number with the underline is the second. }
\centering
\scalebox{0.69}{
\begin{tabular}{c|c|c|c|c|c|c|c|c|c|c|c|c|c}

\hline
 \multicolumn{14}{c}{DDAD \cite{ddad} (Supervised learning)}  
\tabularnewline
\hline

 \multicolumn{7}{c}{Experiment setting}  &  \multicolumn{7}{|c}{Discrete depth metric}
\tabularnewline
\hline

 & $\mathcal{L}_{depth}$  &  $\mathcal{L}_{BCE}$  &  $\mathcal{L}_{L1}$   & Res101  &  LSS \cite{lss} &  Query \cite{wei2023surroundocc} & Abs Rel $\downarrow$ & Sq Rel $\downarrow$ & RMSE $\downarrow$ & RMSE log  $\downarrow$ &  $\delta<1.25$ $\uparrow$ & $\delta<1.25^2$ $\uparrow$ & $\delta<1.25^3$ $\uparrow$
\tabularnewline
\hline

(1) & \Checkmark &  & & & & & 0.208 &	2.684	& 9.510 &	0.339 &	\underline{0.678}	& \underline{0.874}	& \underline{0.938}



\tabularnewline

(2) & & \Checkmark & & & & & 0.221 & 3.204 & 10.283	&  0.396 &	0.668 &	0.849 &	0.914

\tabularnewline

(3) & & & \Checkmark & & & & 0.225	& 3.289 & 10.461 &	0.409 &	0.668 &	0.845 &	0.911

\tabularnewline

(4) & \Checkmark & \Checkmark & & & & & 0.208	 &  \underline{2.654}  & 	9.769 & 	0.349 & 	0.653 & 	0.862	 & 0.933

\tabularnewline

(5) & \Checkmark & & \Checkmark & & & &  0.210 & 2.781 & 9.994 &	0.362 &		0.654	&	0.855 &		0.924

\tabularnewline
\hline

(6) & \Checkmark & & & \Checkmark & & &  \underline{0.206}	& \textbf{2.597} & \textbf{9.353}	& \textbf{0.329}	& \textbf{0.685} &	\textbf{0.878} &	\textbf{0.942}

\tabularnewline

(7) & \Checkmark & & & \Checkmark $^*$ &  &  & 0.265 &	3.933 &	10.356 &	0.369 &	0.620&	0.840&	0.923

\tabularnewline

(8)& \Checkmark & & & \Checkmark & \Checkmark & & 	\textbf{0.205}	& \textbf{2.597}	& \underline{9.470}	& \underline{0.331}	& 0.676	& 0.872 &	\underline{0.938}

\tabularnewline

(9)& \Checkmark & & & \Checkmark & &\Checkmark &  0.208 & 	2.683 &  9.647	&  0.343 &  0.673	&  0.869	& 0.935

\tabularnewline
\hline
-- &  \Checkmark &  \multicolumn{5}{c|}{SurroundOcc \cite{wei2023surroundocc}}  & 0.209	& 2.748 &	9.455 &	0.337 &	0.687	& 0.873	& 0.937

\tabularnewline
\hline

\end{tabular}
}
\vspace{0.2cm}

\label{t:3D occupancy}
\end{table*}

\begin{table*}[t]
\caption{The ablation study of the proposed method in the \textbf{depth metric}. Experiment (7) means do not use the pretrained model from ImageNet. Experiment (6) is the optimal setting, and we regard it as the full model. $\uparrow$ means the value higher is better and $\downarrow$  means lower is better. The number with bold typeface means the best, and the number with the underline is the second. }
\centering
\scalebox{0.69}{
\begin{tabular}{c|c|c|c|c|c|c|c|c|c|c|c|c|c}

\hline
 \multicolumn{14}{c}{DDAD \cite{ddad} (Supervised learning)}  
\tabularnewline
\hline

 \multicolumn{7}{c}{Experiment setting}  &  \multicolumn{7}{|c}{Depth metric}
\tabularnewline
\hline

 & $\mathcal{L}_{depth}$  &  $\mathcal{L}_{BCE}$  &  $\mathcal{L}_{L1}$   & Res101  &  LSS \cite{lss} &  Query \cite{wei2023surroundocc} & Abs Rel $\downarrow$ & Sq Rel $\downarrow$ & RMSE $\downarrow$ & RMSE log  $\downarrow$ &  $\delta<1.25$ $\uparrow$ & $\delta<1.25^2$ $\uparrow$ & $\delta<1.25^3$ $\uparrow$
\tabularnewline
\hline

(1) & \Checkmark &  & & & & & 0.149 &	\underline{1.001} &	4.782 &	\underline{0.222} &	\underline{0.798} &	\underline{0.934} &	\underline{0.972}


\tabularnewline

(2) & & \Checkmark & & & & & 0.175	&1.725	&5.954&	0.289	&0.763&	0.903	&0.949

\tabularnewline
(3) & & & \Checkmark & & & & 0.168&	1.570	&6.580	&0.310	&0.735&	0.882&	0.936

\tabularnewline

(4) & \Checkmark & \Checkmark & & & & & 0.150 &	1.028	&4.868&	0.225	&0.790 &	0.931 &	0.971

\tabularnewline

(5) & \Checkmark & & \Checkmark & & & &  0.151&	1.056	&4.937	&0.230 &	0.790 &	0.929&	0.970

\tabularnewline
\hline

(6) & \Checkmark & & & \Checkmark & & &  \textbf{0.147} &	\textbf{0.996} &	\underline{4.738}  &	\textbf{0.220} &	\textbf{0.801}	& \textbf{0.935} &	\underline{0.972}

\tabularnewline

(7) & \Checkmark & & & \Checkmark $^*$ & & & 0.198	& 1.555	& 5.761	& 0.275	& 0.709	& 0.888	& 0.951

\tabularnewline

(8)& \Checkmark & & & \Checkmark & \Checkmark & & 	\underline{0.148} &	1.013 &	4.797	& \underline{0.222}	& 0.797 &	\underline{0.934} &	\underline{0.972}

\tabularnewline

(9)& \Checkmark & & & \Checkmark & &\Checkmark &  0.150 &	1.008 &	\textbf{4.737} &	\textbf{0.220} &	\underline{0.798} &	\textbf{0.935} &	\textbf{0.973}

\tabularnewline
\hline
-- &  \Checkmark &  \multicolumn{5}{c|}{SurroundOcc \cite{wei2023surroundocc}}  & 0.152	 & 1.022	 & 4.668	 & 0.220	 & 0.797	 & 0.933	 & 0.972

\tabularnewline
\hline

\end{tabular}
}
\vspace{0.2cm}

\label{t:depth metric}
\end{table*}

\begin{table*}[t]
\caption{The ablation study of the proposed method in the proposed discrete depth metric and depth metric. Our network is the optimal Experiment setting (6) in Table \ref{t:3D occupancy}. Our$^{\dagger}$ represents using the PoseNet in \cite{surrounddepth} to predict the 6D pose of two input frames. Our (Density) means the final output is density value without passing the sigmoid function. Our (SDF) means the final output is signed distance value. $\uparrow$ means the value higher is better and $\downarrow$  means lower is better. The number with bold typeface means the best. }
\centering
\scalebox{0.76}{
\begin{tabular}{c|c|c|c|c|c|c|c|c|c|c|c|c|c}

\hline
 \multicolumn{7}{c|}{Method}  & \multicolumn{7}{c}{Nuscenes \cite{nuscenes}}  
\tabularnewline
\hline
 \multicolumn{14}{c}{Supervised learning, Discrete depth metric}
\tabularnewline
\hline

 \multicolumn{7}{c|}{--} & Abs Rel $\downarrow$ & Sq Rel $\downarrow$ & RMSE $\downarrow$ & RMSE log  $\downarrow$ &  $\delta<1.25$ $\uparrow$ & $\delta<1.25^2$ $\uparrow$ & $\delta<1.25^3$ $\uparrow$

 \tabularnewline
\hline

  \multicolumn{7}{c|}{SurroundOcc \cite{wei2023surroundocc}} & 0.361	& 6.089	& 9.726	& 0.391	& 0.585	& 0.820	& 0.923

\tabularnewline
\hline

\multicolumn{7}{c|}{Ours} & \textbf{0.205} & \textbf{2.044} &	\textbf{6.495} &	\textbf{0.289} &	\textbf{0.718} &	\textbf{0.922} &	\textbf{0.966}

\tabularnewline
\hline
\multicolumn{14}{c}{Self-supervised learning, Discrete depth metric}  

\tabularnewline
\hline
 \multicolumn{7}{c|}{SurroundOcc \cite{wei2023surroundocc}} & 0.431	& 8.483 &	12.200 &	0.490 & 0.542	& 0.726 & 0.839

\tabularnewline
\hline
\multicolumn{7}{c|}{Ours$^{\dagger}$} & 0.570	 &  8.835  & 13.453	&  1.152	& 0.211	& 	0.403 & 0.530

\tabularnewline
\hline
\multicolumn{7}{c|}{Ours} &	0.211 &  \textbf{2.460} & 7.687	& 0.384	& 0.702	& 0.855	& 0.914

\tabularnewline
\hline
\multicolumn{7}{c|}{Ours (Density)} & 	0.226	&  3.900	&  7.876	&  0.371 & 	0.736	&  0.871	&  0.927

\tabularnewline
\hline
\multicolumn{7}{c|}{Ours (SDF)} &  \textbf{0.203}	& 2.703 & \textbf{7.534} & \textbf{0.350} &	\textbf{0.732} & \textbf{0.872} & \textbf{0.928}

\tabularnewline
\hline


 \multicolumn{14}{c}{Supervised learning, Depth metric}

 \tabularnewline
\hline

\multicolumn{7}{c|}{SurroundOcc \cite{wei2023surroundocc}} & 0.117 & 0.591	& \textbf{2.623}	& 0.180	& 0.886	& 0.950	& 0.975
\tabularnewline
\hline

\multicolumn{7}{c|}{Ours} & 0.116	& \textbf{0.542}	&  2.784  & \textbf{0.179}  &	\textbf{0.888} & \textbf{0.955} &	\textbf{0.978}

\tabularnewline
\hline
\multicolumn{14}{c}{Self-supervised learning, Depth metric}  

\tabularnewline
\hline
 \multicolumn{7}{c|}{SurroundOcc \cite{wei2023surroundocc}} & 0.274	&  3.405	&  6.127 &  0.345	&  0.705	&  0.854	&  0.920

\tabularnewline
\hline
\multicolumn{7}{c|}{Ours$^{\dagger}$} &	0.725  &   8.174  &  14.013  &   1.344  &   0.011  &   0.024  &   0.043

\tabularnewline
\hline
 \multicolumn{7}{c|}{Ours} & 0.227 &  3.437 &  5.463	 &  0.305	 &  \textbf{0.789}	 &  \textbf{0.899}	 &  0.940
 
\tabularnewline
\hline
\multicolumn{7}{c|}{Ours (Density)} & 	\textbf{0.199}	& \textbf{2.261} &	\textbf{5.396} &	\textbf{0.302}  &	0.775 &	0.893 &	\textbf{0.941}

\tabularnewline
\hline
 \multicolumn{7}{c|}{Ours (SDF)} & 0.210	& 2.655 &	5.540 &	 0.306 &	0.777 & 	0.892  &	 0.939

\tabularnewline
\hline

\end{tabular}
}
\vspace{0.2cm}

\label{t:nuscene 3D occupancy}
\end{table*}

\section{Experiment}
We validate our proposed framework through extensive experiments utilizing the DDAD and Nuscenes datasets. Our experimental evaluation comprises four key components.
Firstly, we conduct a comprehensive ablation study on the proposed framework, delving into the intricacies of the supervised loss function and network architecture.
Subsequently, we delve into the domain of self-supervised learning within the Nuscenes dataset, shedding light on its impact on 3D reconstruction performance.
In the third phase, leveraging volume rendering for depth map acquisition, we establish a benchmark for depth metrics. This entails a comparative analysis between our outcomes and those of monocular depth estimation methods.
Finally, we discuss our work with the recent semantic 3D occupancy estimation method. Our experiments conclusively demonstrate that our point-level optimization approach could benefit semantic 3D occupancy estimation in an effective pretraining strategy.

\subsection{Datasets}

\textbf{DDAD} \cite{ddad} is a largescale dataset with dense ground-truth depth maps. Specifically, this dataset includes 12,650 training samples. The validation set contains 3,950 samples. We only consider the distance up to 52m, which is a reasonable range referred from the 3D semantic completion task \cite{monoscene} and BEV perception task \cite{bevformer}. The DDAD dataset has a denser point cloud compared with Nuscenes dataset, which could provide a more equitable evaluation, so we conduct an ablation study experiment based on the DDAD dataset.

\textbf{Nuscenes} \cite{nuscenes} has 1000 sequences of diverse scenes, where each sequence is approximately 20 seconds in duration. We split the training and testing set the same as the depth estimation task \cite{surrounddepth}, including 20096 samples for training and 6019 samples for validation. 

For the above two datasets, we generate the occupancy and empty space label from the raw point cloud and use the projected depth map 
for training and testing. Specifically, we first define the $Z$, $Y$, and $X$ as depth, height, and width, respectively. For the similar perception range as SemanticKITTI \cite{semantickitti}, the valid point cloud range is set as $Z\in \left ( -52m, 52m \right )$, $Y \in \left ( 0m, 6m \right )$, and $X\in \left ( -52m, 52m \right )$ for the training and testing. With the final voxel resolution in $256\times256\times16$, the size of the resolutions of the grid is (0.41m, 0.41m, 0.38m). To evaluate the representation with occupancy probability, we set the best threshold for obtaining the score of discrete depth metric from the search range from 0 to 1 with an interval of 0.05 in the test set. For the representation with the signed distance function (SDF), we set the best threshold from the search range from -0.5 to 0.5 for the discrete depth metric's score  in the test set. 

\begin{figure*}
    \centering
    \includegraphics[width=1\textwidth]{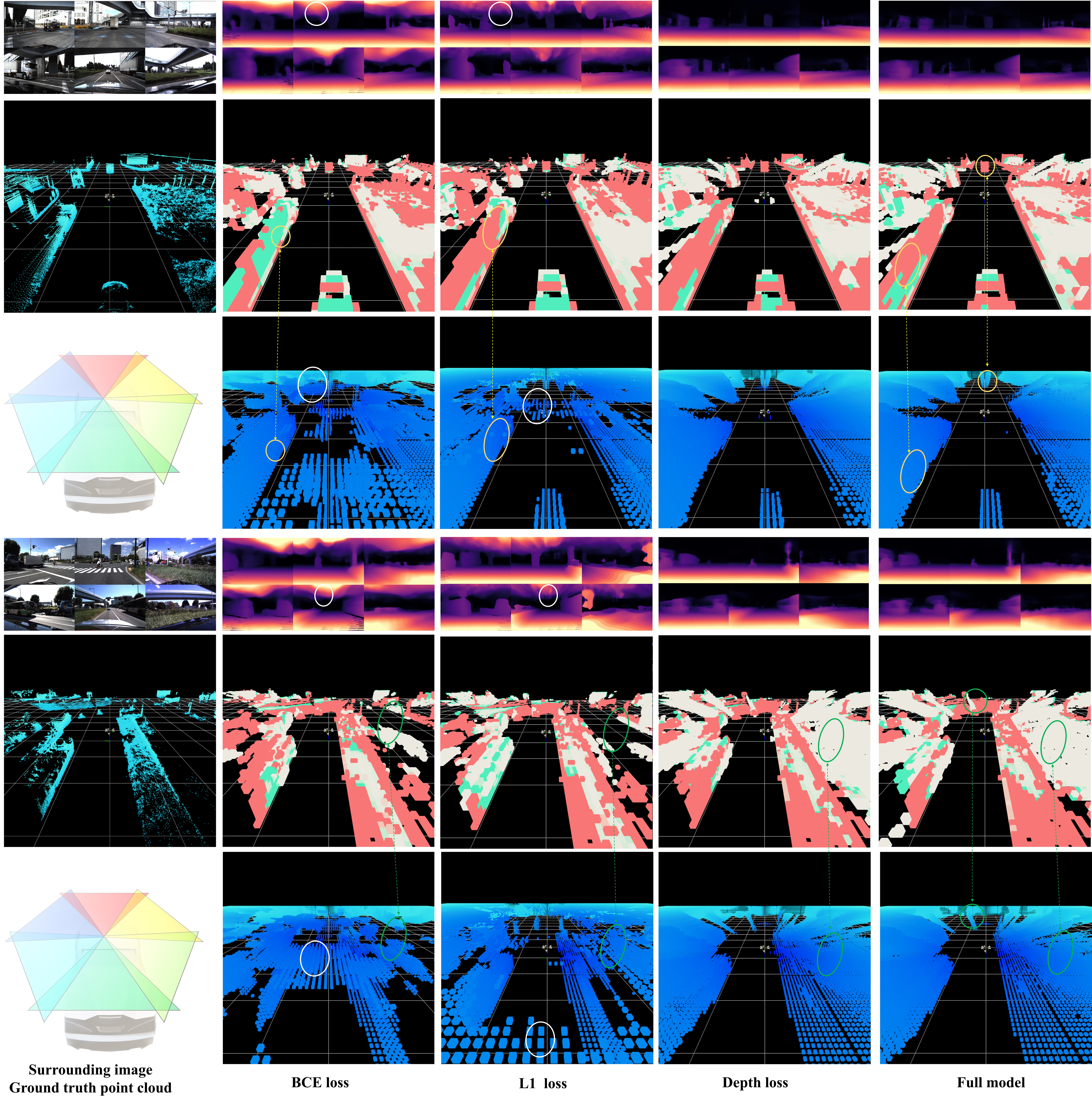}
    \caption{\textbf{The visualization for 3D occupancy and depth estimation ablation study of the proposed method (DDAD dataset \cite{ddad}).} The first row: the surrounding images and the rendered depth maps. For the second row: based on the occupancy label, we present the binary prediction, where the red, green, and white colors mean the false negative, true positive, and false positive, respectively. The third row is the dense occupancy prediction in the voxel grid, where the darker color means the occupancy is closer to the ego vehicle. The BCE loss, L1 loss, Depth loss, and Full model are related to the experiment setting (2), (3), (1), and (6) in Table \ref{t:3D occupancy} and \ref{t:depth metric}, respectively. Note that we omit the prediction under the 0.4 m for better visualization. Best viewed in color.}
    \label{fig:Result_ablation}
\end{figure*}

\begin{figure*}
    \centering
    \includegraphics[width=1\textwidth]{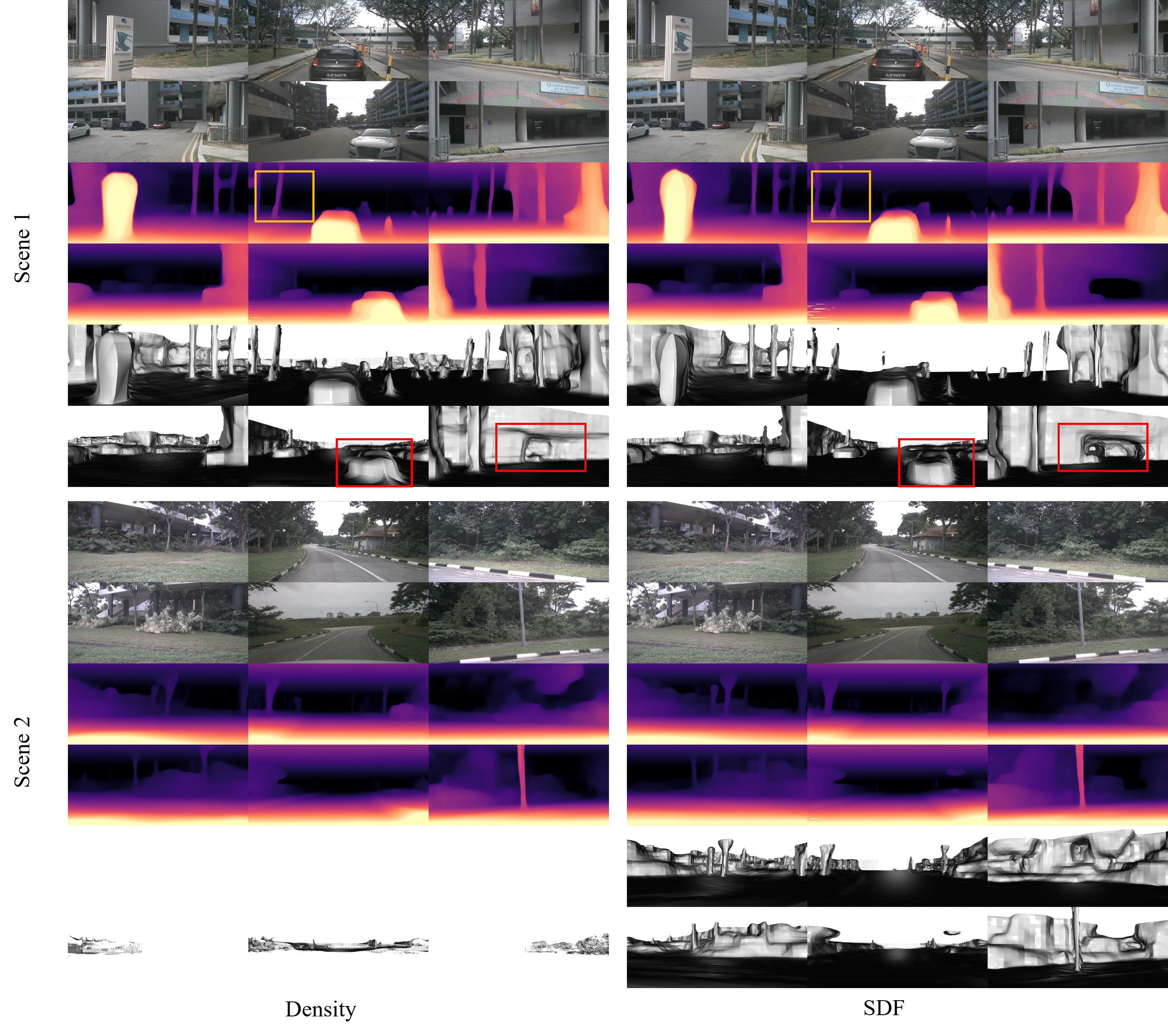}
    \caption{\textbf{The depth map and mesh visualization for comparing the representation of density and signed distance function (SDF) under the self-supervised learning setting, Nuscenes \cite{nuscenes}}. The mesh for density is extracted with the threshold of 0.5. We visualize the mesh at the camera view with the height range from 0 to 5 m. We can learn that the density representation could extract a reasonable mesh for scene 1, but it can not work well in scene 2 in the same threshold. For SDF, we could make a good mesh prediction for both scene 1 and 2.}
    \label{fig:nuscene_ablation}
\end{figure*}

\begin{table*}[t]
\caption{The benchmark in depth map metric with monocular depth estimation methods and recent 3D occupancy network, SurroundOcc \cite{wei2023surroundocc}. Apart from the 52 m, we also extend the perception range up to 80 m for the comparison. SurroundDepth $^{\dagger}$ is the result from the original paper that uses the sparse depth map from structure-from-motion as the scale-aware pertaining, which does not use ground truth pose during the training. The number with bold typeface means the best.}
\centering
\scalebox{0.72}{
\begin{tabular}{l|c|c|c|c|c|c|c|c|c|c}

\hline
Method & Occupancy & Depth & Abs Rel &  Sq Rel &  RMSE &  RMSE log &   $\delta<1.25$  &   $\delta<1.25^{2}$  &  $\delta<1.25^{3}$ & Inference time (s)
\tabularnewline

\hline
 \multicolumn{11}{c}{DDAD \cite{ddad} (Supervised learning, 52 m)}  

\tabularnewline
\hline

Monodepth2 \cite{monodepth2} &  &  \Checkmark &  0.148  &   1.003  &   4.722  &   0.219  &   0.795  &   0.935  &   0.973 &  \textbf{0.029}

\tabularnewline
\hline

New-CRFs \cite{newcrfs} &  &  \Checkmark &  \textbf{0.130}  &  \textbf{0.824}  &   \textbf{4.169}  &   \textbf{0.195}  &   \textbf{0.840}  &   \textbf{0.951}  &   \textbf{0.979} & 0.065

\tabularnewline
\hline

Surrounddepth \cite{surrounddepth} &  &  \Checkmark &   0.151  &   1.021  &   4.766  &   0.221  &   0.789  &   0.935  &   0.974 & 0.099

\tabularnewline
\hline

SurroundOcc \cite{wei2023surroundocc} & \Checkmark &  \Checkmark &  0.152	 & 1.022	 & 4.668	 & 0.220	 & 0.797	 & 0.933	 & 0.972 & 0.305

\tabularnewline
\hline

Ours & \Checkmark &  \Checkmark &  0.147&	0.996	&4.738	& 0.220	& 0.801	& 0.935	& 0.972 & 0.230

\tabularnewline
\hline

 \multicolumn{11}{c}{Nuscenes \cite{nuscenes} (Supervised learning, 52 m)}  
\tabularnewline
\hline

Monodepth2 \cite{monodepth2} &  &  \Checkmark &  0.131  &   0.728  &   3.479  &   0.201  &   0.845  &   0.942  &   0.974 & \textbf{0.024}

\tabularnewline
\hline

New-CRFs \cite{newcrfs} &  &  \Checkmark & \textbf{0.115}  &  0.618  &   3.193  &  0.185  &   0.872  &   0.951  &  0.977  &  0.058

\tabularnewline
\hline

Surrounddepth \cite{surrounddepth} &  &  \Checkmark &  0.128  &   0.760  &   3.442  &   0.198  &   0.856  &   0.946  &   0.975 & 0.102

\tabularnewline
\hline

SurroundOcc \cite{wei2023surroundocc} & \Checkmark &  \Checkmark & 0.117 & 0.591	& \textbf{2.623}	& 0.180	& 0.886	& 0.950	& 0.975 & 0.305

\tabularnewline
\hline

Ours & \Checkmark &  \Checkmark & 0.116	& \textbf{0.542}	&  2.784  & \textbf{0.179}  &	\textbf{0.888} & \textbf{0.955} &	\textbf{0.978} & 0.196

\tabularnewline
\hline

\multicolumn{11}{c}{Nuscenes \cite{nuscenes} (Self-supervised learning, 52 m)}  
\tabularnewline
\hline

Monodepth2 \cite{monodepth2} &  &  \Checkmark &  0.258  &   4.566  &   5.650  &   0.316  &   0.785  &   0.894  &   0.935 & --

\tabularnewline
\hline

New-CRFs \cite{newcrfs} &  &  \Checkmark &  0.230  &   3.931  &   5.361  &   0.304  &   \textbf{0.812}  &   \textbf{0.904}  &   0.939  & --

\tabularnewline
\hline

Surrounddepth \cite{surrounddepth} &  &  \Checkmark & 0.215  &   3.127  &   \textbf{5.277}  &   \textbf{0.294}  &   0.797  &   0.903  &   \textbf{0.943} &  --

\tabularnewline
\hline

SurroundOcc \cite{wei2023surroundocc} & \Checkmark &  \Checkmark &  0.274  &   3.405  &   6.127  &   0.345  &   0.705  &   0.854  &   0.920 & --
\tabularnewline
\hline


Ours & \Checkmark &  \Checkmark &  \textbf{0.199}	& \textbf{2.261} &	5.396 &	0.302  &	0.775 &	0.893 &	0.941  & --
\tabularnewline
\hline

\multicolumn{11}{c}{Nuscenes \cite{nuscenes} (Self-supervised learning, 80 m)}  
\tabularnewline
\hline









Monodepth2 \cite{monodepth2} &  &  \Checkmark &  0.332  &  10.809  &   7.907  &   0.352  &   0.776  &   0.883  &   0.926  & --

\tabularnewline
\hline

New-CRFs \cite{newcrfs} &  &  \Checkmark & 0.295  &   9.358  &   7.571  &   0.339  &   \textbf{0.802}  &   \textbf{0.894}  &   \textbf{0.930} & --

\tabularnewline
\hline

Surrounddepth \cite{surrounddepth} &  &  \Checkmark &  0.320  &  10.811  &   7.936  &   0.345  &   0.783  &   0.889  &   \textbf{0.930}  & --

\tabularnewline
\hline

SurroundDepth $^{\dagger}$ \cite{surrounddepth}  &  &  \Checkmark    & 0.280 & 4.401 & 7.467 & 0.364 & 0.661 & 0.844 & 0.917   & --

\tabularnewline
\hline

SurroundOcc \cite{wei2023surroundocc} & \Checkmark &  \Checkmark &  0.488  &   7.072  &   8.979  &   0.458  &   0.571  &   0.749  &   0.847 & --

\tabularnewline
\hline

Ours & \Checkmark &  \Checkmark & \textbf{0.224}  &   \textbf{3.383}  &   \textbf{7.165}  &   \textbf{0.333}  &   0.753  &   0.877  &   \textbf{0.930}  & --

\tabularnewline
\hline

\end{tabular}
}
\vspace{0.3cm}

\label{t:depth2}
\end{table*}

\subsection{Implementation detail}
Our approach is implemented with Pytorch \cite{pytorch}. We resize the RGB image into $336\times672$ before putting them into the neural network. The depth map is rendered with the resolution $224\times352$. We use the Adam optimizer \cite{adam} with $\beta 1 = 0.9$ and $\beta 2 = 0.999$. The learning rate of the 2D CNN is set as $1e-4$ and the 3D CNN is $1e-3$. We train the networks with 12 epochs and do the learning rate decay with 0.1 at the 10th epoch. All the experiments have been conducted on the NVIDIA A 100 (40 GB) GPU.

\subsection{The ablation study for supervised loss and network architecture}

We present the ablation study result of the proposed method in Table \ref{t:3D occupancy} (discrete depth metric) and Table \ref{t:depth metric} (depth metric). To conserve space, we also include the experimental results for SurroundOcc \cite{wei2023surroundocc} in the table for a direct comparison with our proposed optimal network under the specified experimental setting (6).
We investigate the characteristics of the different supervised loss functions and the network design as follows:

\textbf{Supervised loss functions analysis.} First, we conduct the experiments (1-5) with different loss functions based on the lighter encoder, Resnet 50. We can see that the depth loss ($\mathcal{L}_{depth}$) outperforms the classification losses ($\mathcal{L}_{BCE}$ and $\mathcal{L}_{L1}$) for both the discrete depth metric and the depth metric in Table \ref{t:3D occupancy} and Table \ref{t:depth metric}, in the experiment setting (1-3). Accompanying with Figure \ref{fig:Result_ablation} marked with white circles in the depth map and the dense occupancy map, we can observe that the classification losses easily produce the floater in the region close to the sky. We conclude that the training with classification loss could not handle these regions behind the point cloud (unknown region), and this situation is also observed in the following pretrain experiment as shown in Figure \ref{fig:pretrain}. Reversely, the depth loss could largely prevent this situation because the rendering fashion cloud implicitly optimizes these regions with the sampled points along the whole ray. On the other hand, the depth loss from rendering more easily produced the long tail false positive prediction, as marked with the green circles. The long tail false positive prediction usually happens at the foreground and background intersection, which is not in the concerned driving region. We further investigate the combination of depth loss and classification loss in experiments (4) and (5). However, the combined loss is a little worse than the depth loss. It may own that the classification loss hurts the prediction in the sky region, as analyzed above.

In terms of the metric design, the prediction case 2 in Figure \ref{fig:label} usually happens as marked with yellow circles, where the network fails to predict the first occupancy at the end of the point cloud, but in the real situation, the first occupied point is very close to the point cloud from the visualization of the dense occupancy. Therefore, based on the quantitative analysis in Figure \ref{fig:label} and the qualitative observation in Figure \ref{fig:Result_ablation}, the proposed discrete depth metric could have a more justice assessment for the 3D occupancy estimation.

\textbf{Network design.} Building upon the depth loss, we delve into the investigation of network design. Note that in this section, our focus is not on introducing a highly specific network component but rather on exploring the performance of existing modules within the context of BEV perception under our proposed framework. At first, we try with a larger encoder by replacing ResNet 50 with ResNet 101, and the model performs better. Note that, in the experiment (7), the model gets a severe performance drop without initializing the model pre-trained from on ImageNet, which hints that finding a better pretrain strategy associated with 3D occupancy estimation may further boost the performance. The used parameter-free interpolation to recover 3D feature space is the most straightforward and efficient manner. In experiment (8), we also try to use the unprojection manner proposed in LSS \cite{lss}, which estimates the depth distribution first, and then forms the 3D volume with the weighted feature based on the distribution information. However, we overlooked the benefit of LSS's unprojection manner on the 3D occupancy estimation task, which gives us the information that doing the depth distribution estimation may cause a larger learning space for the following 3D CNN feature aggregation module, especially under the imperfect depth distribution estimation. Similarly, we have not observed the performance gain with the try on the Query method \cite{wei2023surroundocc}, which is based on the transformer with cross attention. 




\subsection{Self-supervised learning and 3D reconstruction}

For our investigation into self-supervised learning and 3D reconstruction, we perform experiments on the Nuscenes dataset \cite{nuscenes}. This choice was motivated by the higher accuracy of the provided 6D pose for the two subsequent frameworks, as opposed to the DDAD dataset \cite{ddad}. The results of this exploration are summarized in Table \ref{t:nuscene 3D occupancy}, where we also include results from supervised learning for comparison. This supervised setting mirrors the conditions of experiment (6) in Table \ref{t:3D occupancy}.

\begin{figure}
    \centering
    \includegraphics[width=\linewidth]{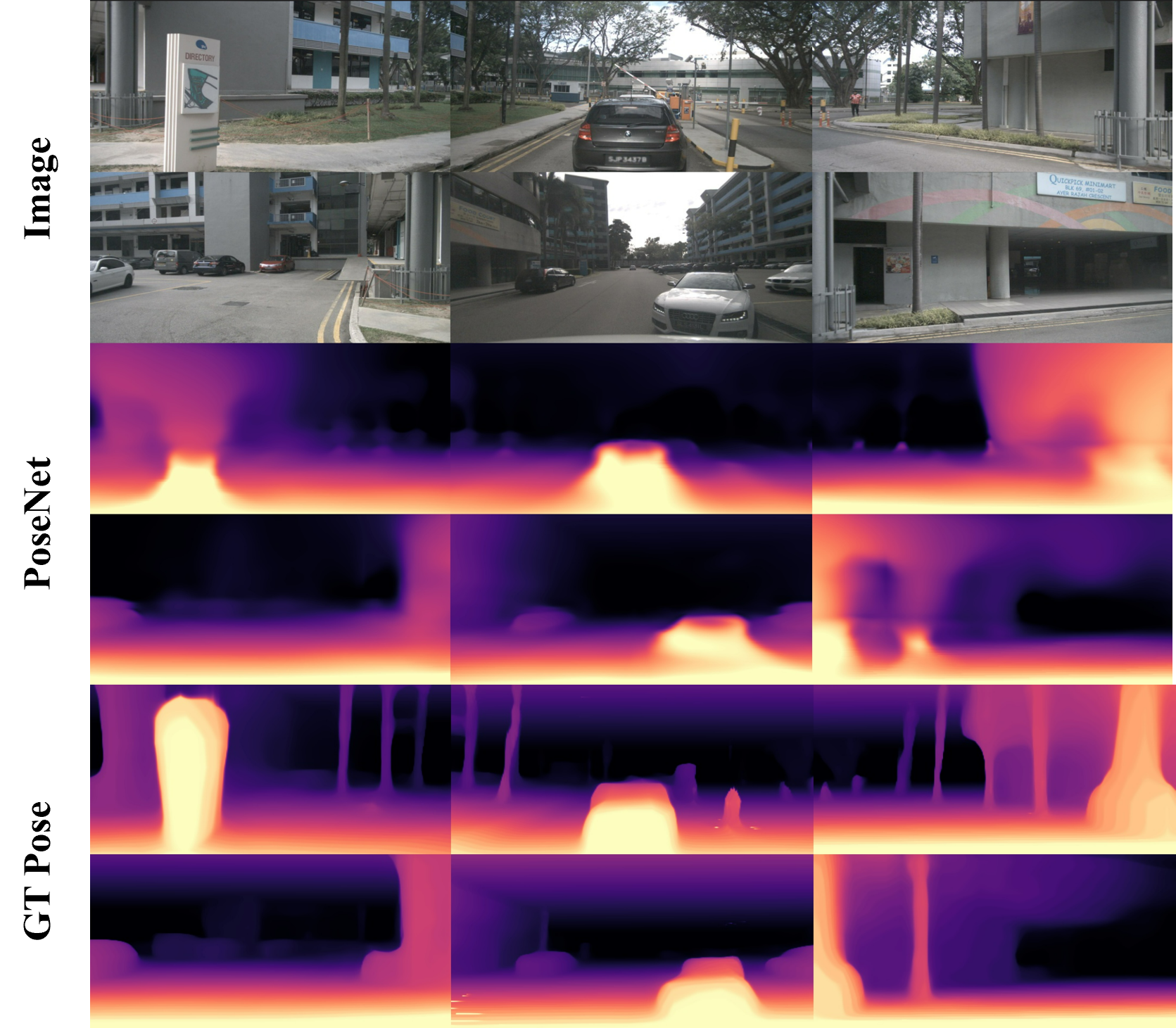}
    \caption{\textbf{The depth map comparison with the pose from PoseNet and ground truth (GT) pose.} The joint training with PoseNet would lead to sub-optimal results due to the scale ambiguity between the predicted depth map and the translation in the predicted 6D pose.}
    \label{fig:scale_ambiguity}
\end{figure}

Our initial observation reveals a persistent gap between supervised and self-supervised learning, even when employing ground truth pose information from the sensor. Notably, as shown in Figure \ref{fig:scale_ambiguity}, the use of pose from PoseNet yields notably inferior results due to the inherent scale ambiguity between depth estimation and 6D pose transformation. 
We then delved into the final output representation in the 3D volume space. Table \ref{t:nuscene 3D occupancy} highlights that the Signed Distance Function (SDF) representation outperforms in discrete depth metrics, indicating its superiority in 3D reconstruction, as also evidenced by the marked improvement in the region enclosed by the red rectangle in Figure \ref{fig:nuscene_ablation}. Conversely, the SDF representation yields a less favorable result in the depth metric, as demonstrated in Table \ref{t:nuscene 3D occupancy} and visually represented by the yellow rectangle in Figure \ref{fig:nuscene_ablation}. This discrepancy can be attributed to the relatively challenging nature of optimizing signed distance values, while the density and probability representations offer more flexibility for optimization.
However, it's worth noting that this soft and flexible characteristic is less conducive to mesh extraction at specific thresholds. This is exemplified in Figure \ref{fig:nuscene_ablation}, where the mesh output in scene 1 is reasonable for density representation, but the result in scene 2 is suboptimal. Consequently, for the specific purpose of 3D reconstruction, the SDF representation emerges as the superior choice.

\subsection{The benchmark for the depth estimation}
In the surrounding-view setting, we build a new benchmark in terms of the depth map metric for comparison with the monocular depth estimation methods. Monodepth2 \cite{monodepth2} is a well-recognized self-supervised monocular depth estimation method and New-CRFs \cite{newcrfs} is the state-of-the-art supervised method. Surrounddepth \cite{surrounddepth} is a self-supervised work that has been introduced in the related work. For the supervised learning in this work, we train the above three monocular depth estimation methods with the same loss function used in \cite{newcrfs}. For self-supervised learning, we use the loss function in Equation \ref{self-supervised loss} for all the compared methods.

\begin{figure*}
    \centering
    \includegraphics[width=0.85\textwidth]{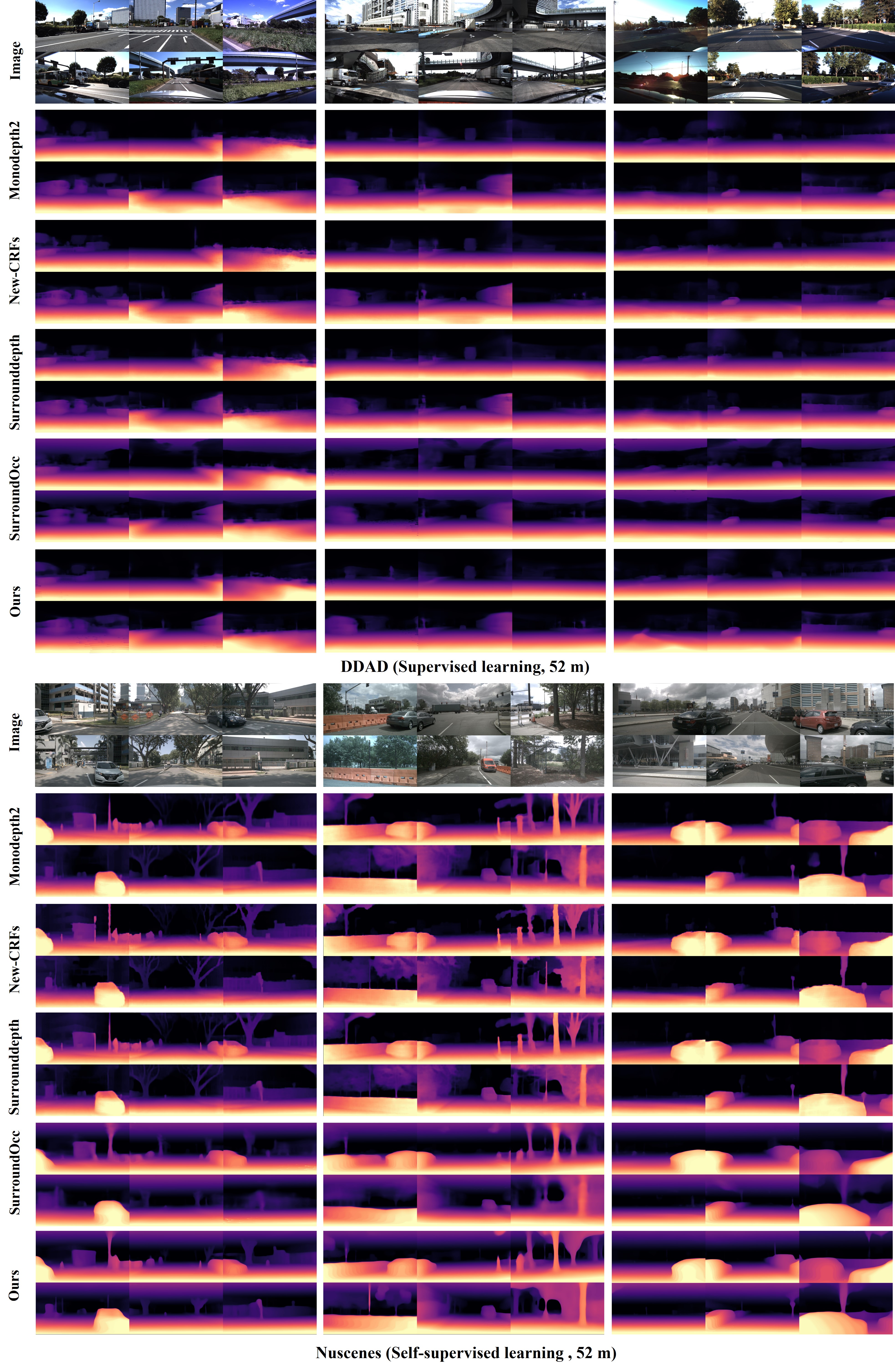}
    \caption{\textbf{The depth map visualization results for DDAD dataset in the supervised learning and Nuscenes dataset in the self-supervised learning.} For the camera's order in the surrounding images and the rendered depth maps, the first row is front left, front and front right, and the second row is back left, back, and back right. Best viewed in color.}
    \label{fig:DDAD_Nuscenes}
\end{figure*}

From Table \ref{t:depth2}, we can observe that our method is competitive with the monocular depth estimation methods. For the DDAD dataset, New-CRFs \cite{newcrfs} achieves the best result, matching its performance in single image depth estimation. The performance for Monodepth2 \cite{monodepth2}, Surrounddepth \cite{surrounddepth}, and ours is similar. For the Nuscenes dataset, our method achieves the best result in supervised learning and competitive performance in self-supervised learning. In particular, for self-supervised learning, we observe that our method generally achieves better results for the error metrics and a little worse than the monocular depth estimation methods in terms of the accuracy metrics, which may own to the discrepancy between volume rendering (ours) and encoder-decoder (monocular depth estimation) fashion. In Figure \ref{fig:DDAD_Nuscenes}, We present the depth map visualization on the DDAD dataset in the supervised learning and the visualization result for Nuscenes dataset in the self-supervised setting. More visualization result is presented in the supplementary material. The sequence demonstrations for the 3D occupancy, depth map, and mesh have been attached to the code page \footnote{\href{https://github.com/GANWANSHUI/SimpleOccupancy}{https://github.com/GANWANSHUI/SimpleOccupancy}}.

The disadvantage of our method is that the inference time is higher than the monocular depth estimation methods. We further analyze the inference time for each component of the system. For the DDAD dataset: 2D CNN (31) + 2D to 3D interpolation (3) + 3D CNN (101) + Rendering (95) = 230 (ms). We learn that the 3D feature aggregation and the rendering of the depth map occupied the main time. Note that Rendering is for depth maps and is ignorable if we only want the occupancy estimation.

\subsection{Discussion on semantic 3D occupancy estimation}

\textbf{SurroundOcc in our framework.} The motivation behind our approach to 3D occupancy estimation aligns with concurrent works \cite{wei2023surroundocc,wang2023openoccupancy, tong2023scene} in the field. Among them, we select SurroundOcc \cite{wei2023surroundocc} as a representative for discussion limited by the computational cost. First, We implement the same loss function to SurroundOcc with the similar training strategy, where the detailed configuration is in the supplemental material. From Table \ref{t:3D occupancy} to Table \ref{t:depth metric}, we could find that our network is performing better than it for both the performance and computational cost. Note that, during the experiment, we observe that SurroundOcc is relatively difficult to optimize in self-supervised learning and easily falls into the sub-optimal results, hence, with worse results. 

\begin{figure}
    \centering
    \includegraphics[width=\linewidth]{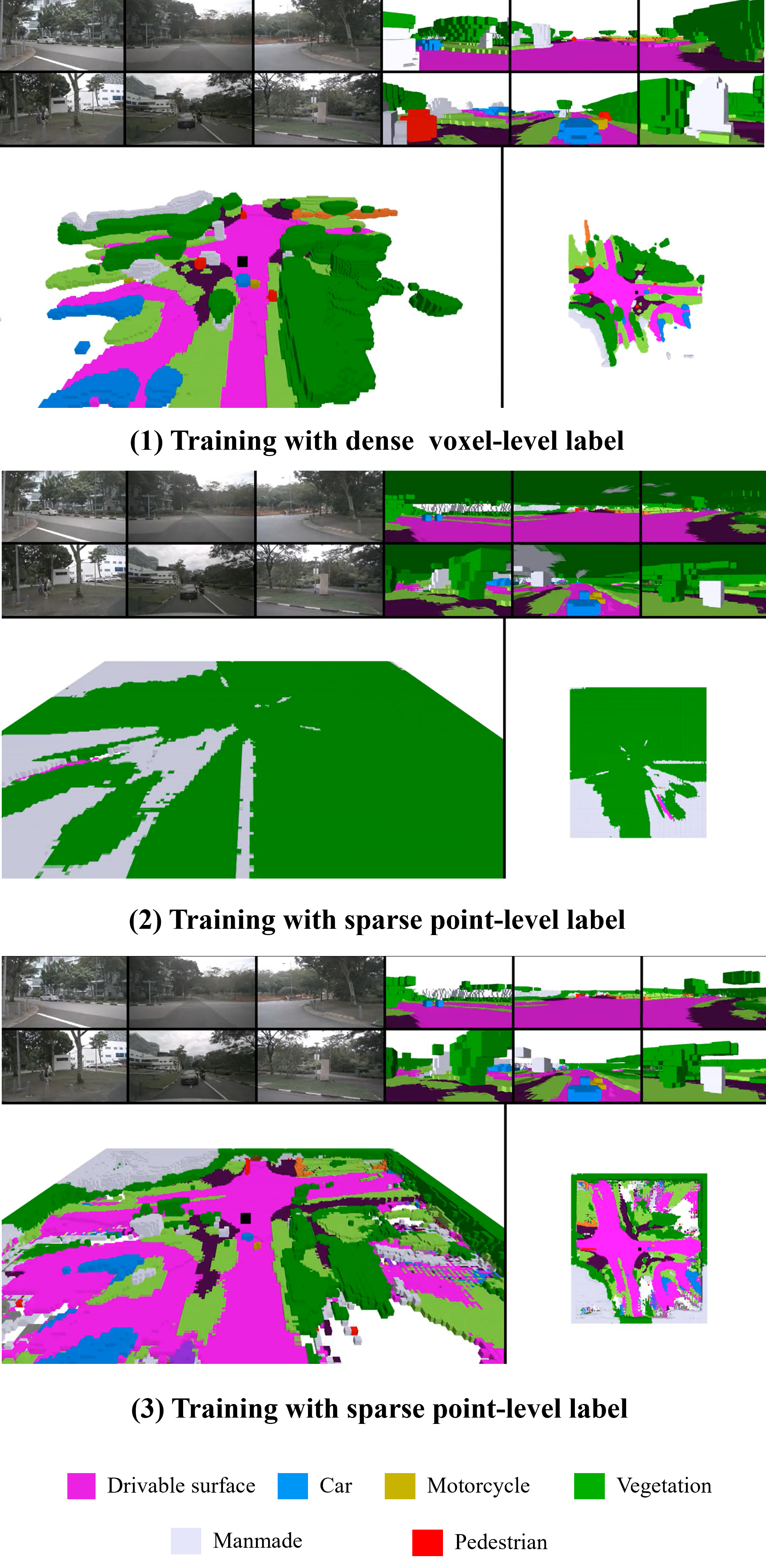}
    \caption{\textbf{The comparison for the (1) dense voxel-level supervision from SurroundOcc \cite{wei2023surroundocc} and (2,3) point-level supervision by our sampling strategy directly from the point cloud label. (3) is the result after removing the top prediction of (2), where we can see that the point-level optimization could produce a reasonable prediction without the need for dense voxel label.} }
    \label{fig:pretrain}
\end{figure}

\textbf{Pretrain strategy for SurroundOcc.} Second, as depicted in Figure \ref{fig:label}, our method adopts a point-level training approach, while the concurrent works employ voxel-level training. This distinction allows us to achieve a finer level of granularity in our predictions compared to theirs. To explore the potential benefits of our strategy for concurrent work, we conduct an experiment using our approach as a pretraining step. Specifically, we train SurroundOcc on point-level semantic labels generated by our sampling strategy, and then we fine-tune the network using voxel-level semantic labels from SurroundOcc. More implementation details are placed in the supplement. The results in Table \ref{t:pretrain} demonstrate that our point-level pretraining effectively improves the network's performance. As shown in Figure \ref{fig:pretrain}, the outcome of training with sparse point-level labels is not optimal. In the upper part of the view, we observe that vegetation and manmade structures dominate, as these classes typically appear near the top. Consequently, the network tends to predict all regions close to the sky as either vegetation or manmade. Despite this limitation, the network demonstrates reasonable predictions for the remaining parts of the scene, which proves to be a valuable initialization for subsequent fine-tuning and leads to improved performance.

\section{Limitation and future work}
In this simple framework for 3D occupancy estimation, we still do not introduce sequence information as did in \cite{tesla} and the BEV perception tasks \cite{bevformer, liu2022petrv2}. It is a promising direction to improve performance by fusing the sequence information. Besides, the current voxel resolution is relatively coarse with 0.4 m, which is a good beginning for the research purpose due to limited computation resources. In the future, we will explore a higher resolution, such as 0.2 m in \cite{monoscene}, which should also benefit the 3D reconstruction. In additional, in this framework, we directly output the geometry attribute in the final 3D voxel space which is constrained by the computation cost. To further improve the 3D geometry representation, a solution is to use the MLPs to regress the final geometry attribute so that the additional regularization loss such as eikonal loss \cite{gropp2020implicit, li2023neuralangelo} could be implemented.

\begin{table*}[t]
\caption{The ablation study on SurroundOcc with point level pretrain. Original is from the original paper. W/O pretrain represents our reimplement result. W/ pretrain means using our point level pretrain. The number with bold typeface means the best.}
\centering
\scalebox{0.78}{
\begin{tabular}{c|c|c|c}

\hline
 Method & Experiment setting  &  SC IoU  &  SSC IoU
\tabularnewline
\hline

 \multirow{3}*{SurroundOcc \cite{wei2023surroundocc}} & Original   & 31.49	 & 20.30

\tabularnewline

~ & W/O pretrain   &	31.29   & 20.08

\tabularnewline

~ & W pretrain   &	\textbf{32.55}  &  \textbf{20.85}

\tabularnewline
\hline


\end{tabular}}

\vspace{0.3cm}
\label{t:pretrain}
\end{table*}

\section{Conclusion}
In this paper, we give a simple framework for 3D occupancy estimation, depth estimation, and 3D reconstruction in the surrounding-view setting for autonomous driving. We demonstrate the effectiveness of the entire pipeline through novel network design, loss function investigation, and model evaluation. Besides, we reveal a pretrain strategy for 3D semantic occupancy prediction based on our sampling strategy. 
We hope this simple and effective framework will inspire followers, and we have released the code to encourage further research in this field.










\bibliographystyle{ieeenat_fullname}
\bibliography{sn-bibliography}




\end{document}


\title[Article Title]{Supplementary material}


\author[1,2]{\fnm{Wanshui} \sur{GAN}}\email{wanshuigan@gmail.com}

\author[3]{\fnm{Ningkai} \sur{Mo}}\email{nk.mo19941001@gmail.com}

\author[4]{\fnm{Hongbin} \sur{Xu}}\email{hongbinxu1013@gmail.com}

\author[1,2]{\fnm{Naoto} \sur{Yokoya}}\email{yokoya@k.u-tokyo.ac.jp}
\equalcont{corresponding author.}

\affil[1]{\orgname{The University of Tokyo}, \orgaddress{ \state{Tokyo}, \country{Japan}}}

\affil[2]{\orgname{RIKEN }, \orgaddress{\state{Tokyo}, \country{Japan}}}

\affil[3]{\orgname{Shenzhen Institute of Advanced Technology, Chinese Academy of Sciences}, \orgaddress{ \state{Shenzhen}, \country{China}}}

\affil[4]{\orgname{South China University of Technology}, \orgaddress{\state{Guangzhou}, \country{China}}}



\abstract{
In this supplementary material, we provide more implementation details, experimental results, and analysis.}

\maketitle

\begin{appendices}





\section{More implementation details}

\textbf{The detailed architecture of the 3D CNN.} For the volume feature aggregation, we adapt the 3D CNN from the HybirdNet \cite{HybirdNet} as shown in Figure \ref{fig:3DCNN}. The difference is that we discard the multiple 3D volume output from the stacked hourglass architecture because we do not find the performance improvement with the default architecture, and rendering for the multiple depth maps also need more computational resource. Please find the open-source code for a more detailed design of the 3D CNN. 

\textbf{Depth map benchmark.} To build up the depth map benchmark, we implement the same training strategy for Monodepth2 \cite{monodepth2}, New-CRFs \cite{newcrfs}, and Surrounddepth \cite{surrounddepth}. Specifically, we train the networks with 15 epochs and do the learning rate decay with 0.1 at the 12th epoch for the DDAD dataset. For the Nuscenes dataset, we train the networks with 10 epochs and do the learning rate decay with 0.1 at the 8th epoch as this dataset is faster to converge, which is also observed in Surrounddepth \cite{surrounddepth}. 

\textbf{The detailed implementation on SurroundOcc \cite{wei2023surroundocc}.}
For the experiment of SurroundOcc, we made the following adjustments:
1. We set the final channel to 1 instead of 18, as our focus is on pure geometry prediction without considering semantic information.
2. To ensure a fair comparison, we set the final output size to $256\times256\times16$, which is the same as ours.
3. We used the same learning rate as ours, as we found that the original learning rate in the reference paper led to inferior results.

Regarding the pretrain experiment (Table 4 in the main paper), we followed these steps:
1. We generated point-level labels by sampling from the point cloud with semantic labels provided from the dataset \cite{fong2022panoptic} used by SurroundOCC. The sampling process is depicted in Figure 3 in the main paper, where the sampled points (except the point cloud itself) are assigned to the empty category.
2. We performed classification training based on the generated point-level labels for the pretraining phase.
3. Once the pretraining was completed, we finetuned the network using dense voxel labels to obtain the final scores.

\textbf{Detailed definition of the classification metric.} By representing the scenes with a set of labeled key points, we can use the classification metric to evaluate the model, where we define the prediction with: True Positive $(TP)$, False Positive $(FP)$, True Negative $(TN)$, False Negative $(FN)$. Following the common practice, we report the F1 score $((2\times P \times R)/(P+R))$, Precision $(P = TP/(TP+FP))$, Recall $(R=TP/(TP+FN))$ and Accuracy $((TP+TN)/(TP+FP+TN+FN))$. The IOU (Intersection-Over-Union) metric is calculated from the average IOU $(TP / (TP + FN + FP))$ of each class.

\textbf{Detailed definition of (discrete) depth map metric.} Following the depth estimation task \cite{surrounddepth}, we report the (discrete) depth map evaluation with the following metrics, 
\begin{equation}
\begin{split}
\text {Abs Rel:} \frac{1}{|M|} \sum_{d \in M}\left|\hat{d}-d^*\right| / d^* , \\ 
\text {Sq Rel:} \frac{1}{|M|} \sum_{d \in M}\left\|\hat{d}-d^*\right\|^2 / d^*, \\
\text { RMSE: } \sqrt{\frac{1}{|M|} \sum_{d \in M}|| \hat{d}-d^* \|^2}, \\
\text { RMSE log: } \sqrt{\frac{1}{|M|} \sum_{d \in M}|| \log \hat{d}-\log d^* \|^2} , \\
\delta<t: \% \text { of } d \text { s.t. } \max \left(\frac{\hat{d}}{d^*}, \frac{d^*}{\hat{d}}\right)=\delta<t,
\end{split}
\end{equation}
where the definition of symbols is identical to that of the main paper.

\begin{table*}[t]
\centering
\scalebox{0.78}{
\begin{tabular}{l|c|c|c|c|c|c|c}

\hline
 \multicolumn{8}{c}{DDAD \cite{ddad}}  
\tabularnewline
\hline

 \multicolumn{2}{c}{Experiment setting}  &  \multicolumn{3}{|c}{Discrete depth metric}  & \multicolumn{3}{|c}{Depth metric}
\tabularnewline
\hline

    &  & Abs Rel $\downarrow$ & Sq Rel $\downarrow$ & $\delta<1.25$ $\uparrow$ & Abs Rel $\downarrow$ & Sq Rel $\downarrow$ & $\delta<1.25$ $\uparrow$ 
\tabularnewline
\hline

\multirow{2}*{Two channels} & $\mathcal{L}_{BCE}$  & 0.375 & 8.327 & 0.628  & 0.172 & 1.658 & 0.723

\tabularnewline
~ & $\mathcal{L}_{depth}$  & 0.295 &	5.422 & 0.629 &  \textbf{0.151}  & 1.055 & 0.790

\tabularnewline
\hline
 \multirow{2}*{Single channel} & $\mathcal{L}_{BCE} (\omega = 0.5)$ & 0.379 & 6.383 & 0.626 &  0.176 & 1.699 & 0.710 

\tabularnewline

~ & $\mathcal{L}_{depth} (\omega = 0.5)$  &  0.305	 &  5.629	& 0.617 &  0.152 &  \textbf{1.034} & \textbf{0.795}

\tabularnewline
\hline

 \multirow{2}*{Single channel}  & $\mathcal{L}_{BCE} (\omega = 0.2)$ & 	0.235 &	3.395  & 0.647 & 0.176 & 1.699 & 0.710 

\tabularnewline

~ & $\mathcal{L}_{depth} (\omega = 0.05)$  &  \textbf{0.210} &	\textbf{2.724} & \textbf{0.674}  & 0.152 &  \textbf{1.034} & \textbf{0.795}

\tabularnewline
\hline

\end{tabular}
}
\vspace{0.3cm}
\caption{The comparison of the output channel for the final occupancy prediction. $\omega$ is the threshold for the occupancy prediction, where the probability value is larger than the $\omega$ would be regarded as occupancy. For the single-channel setting, different thresholds have different occupancy predictions, resulting in different discrete depth values. The depth metric is identical to different thresholds since it is not determined by a certain threshold. The number with bold typeface means the best.}
\label{t:output channel}
\end{table*}

\begin{figure*}
    \centering
    \includegraphics[width=1\textwidth]{fig/3DCNN_new.png}
    \caption{\textbf{The architecture of the proposed 3D CNN feature aggregation module.}}
    \label{fig:3DCNN}
\end{figure*}

\begin{figure*}
    \centering
    \includegraphics[width=0.9\textwidth]{fig/channel.jpg}
    \caption{\textbf{The comparison of the output channel for the final occupancy prediction.} The upper part is the comparison of binary cross entropy (BCE) loss. The lower part is the comparison of the depth loss. }
    \label{fig:channelcomparison}
\end{figure*}

\section{More experiment results}

\textbf{The comparison of the output channel for the final occupancy prediction.} In this work, the occupancy prediction can be obtained in two ways. The first approach (single channel), presented in the main paper, utilizes the Sigmoid function and a threshold to determine whether the sampled point is occupied. The second approach involves using two channels for the final output and applying Softmax and Argmax operations to obtain the occupancy prediction. Regarding the two channels setting, we can still render the depth map by selecting the value of the occupancy channel after the Softmax function. Our experiment found that the first approach achieves better results in the discrete depth metric and performs similarly in the depth metric, as indicated in Table \ref{t:output channel}. We observe the results are similar if set the single channel's threshold $\omega  = 0.5$ compared with the two-channel setting, which means the single-channel and two-channel settings are equivalent. Meanwhile, in the single channel setting, it would produce more false negative predictions with a higher threshold. Therefore, we could select a suitable threshold for better occupancy performance in discrete depth metrics. We visualize a scene in Figure \ref{fig:channelcomparison} for both BCE loss and depth loss for reference. Note that this experiment is from our relatively early experiment, and we slightly adjust the network, so the numerical result and visualization are slightly different from the main paper. It still gives you the information for the comparison of output channels.

\textbf{More visualization results for DDAD \cite{ddad} and Nuscenes \cite{nuscenes} datasets.} More 3D occupancy visualization results for DDAD dataset are presented in Figure \ref{fig:ddad}, and the results for Nuscenes are presented in Figure \ref{fig:nuscenes}. We can learn that the model could predict the detailed layout of the surrounding scenes. We hope this baseline work could inspire researchers to further develop this relatively new task. 

\textbf{Depth map visualization for the depth estimation benchmark in Nuscenes.} The depth map visualization results for the Nuscenes dataset are shown in Figure \ref{fig:nuscenes_depth} with the supervised learning setting and in Figure \ref{fig:nuscenes_depth_self_supervised} with the self-supervised learning setting. Note that the depth map results in the Nuscenes dataset with the supervised learning setting have some wave-like pattern due to the point cloud's sparseness. The result in the DDAD dataset does not have this problem with a more dense point cloud for training. Besides, the result with the self-supervised learning setting also does not have this problem since the photometric consistency loss also has a dense supervision signal. The combination of supervised loss and self-supervised loss is interesting for investigation in the future. The quantitative and qualitative results show that the depth map from volume rendering is competitive with the current well-developed monocular depth estimation methods. Besides, the chosen rendering pipeline can be easily extended to incorporate sequence fusion in the future. This would help address the current ill-posed problem setting of inferring 3D geometry from a single image, which is a limitation of current monocular depth estimation methods.

\textbf{Analysis for the Chamfer distance and the proposed discrete distance metric.}
The Chamfer distance is widely used for the evaluation in 3D reconstruction, which is based on finding the closest distance between the prediction and the ground truth (GT). As illustrated in Figure \ref{fig:chamfer}, we find that the chamfer distance is not a good option for the evaluation of autonomous driving. The closest voxel is easy to be the plane ground rather than the GT vehicle. Therefore, the proposed discrete depth metric is a better choice, which is an ego-centric evaluation.

\begin{figure*}
    \centering
    \includegraphics[width=0.85\textwidth]{fig/Chamfer.png}
    \caption{\textbf{The analysis for the Chamfer distance and the proposed discrete distance metric.}  The closest voxel is easy to be the plane ground rather than the GT vehicle. Therefore, the proposed discrete depth metric is a better choice, which is an ego-centric evaluation.
    }
    \label{fig:chamfer}
\end{figure*}

\begin{figure*}
    \centering
    \includegraphics[width=0.92\textwidth]{fig/ddad_new_1.jpg}
    \caption{\textbf{More 3D occupancy visualization results for DDAD dataset.} The result is under the full model (experiment setting (6)). For the third column, we present the binary prediction based on the generated labels, where the red, green, and white colors mean the false negative, true positive, and false positive, respectively. The fourth column is the dense occupancy prediction in the voxel grid, where the darker color means the occupancy is closer to the ego vehicle. Note that we omit the prediction under 0.4 m for better visualization. Best viewed in color.}
    \label{fig:ddad}
\end{figure*}

\begin{figure*}
    \centering
    \includegraphics[width=0.92\textwidth]{fig/nuscene_new_1.jpg}
   \caption{\textbf{The 3D occupancy visualization results for Nuscenes dataset.} The result is under the full model (experiment setting (6)). For the third column, we present the binary prediction based on the generated labels, where the red, green, and white colors mean the false negative, true positive, and false positive, respectively. The fourth column is the dense occupancy prediction in the voxel grid, where the darker color means the occupancy is closer to the ego vehicle. Note that we omit the prediction under 0.4 m for better visualization. Best viewed in color.}
    \label{fig:nuscenes}
\end{figure*}

\begin{figure*}
    \centering
    \includegraphics[width=0.85\textwidth]{fig/nuscene_depth_new_1.jpg}
    \caption{\textbf{The depth map visualization results for Nuscenes dataset in the supervised setting.} For the camera's order in the surrounding images and the rendered depth maps, the first row is front left, front and front right, and the second row is back left, back, and back right. Best viewed in color.}
    \label{fig:nuscenes_depth}
\end{figure*}

\begin{figure*}
    \centering
    \includegraphics[width=0.85\textwidth]{fig/nuscene_depth_self_supervised.jpg}
    \caption{\textbf{The depth map visualization results for Nuscenes dataset in the self-supervised setting.} For the camera's order in the surrounding images and the rendered depth maps, the first row is front left, front and front right, and the second row is back left, back, and back right. Best viewed in color.}
    \label{fig:nuscenes_depth_self_supervised}
\end{figure*}

\end{appendices}

\clearpage

\bibliographystyle{ieeenat_fullname}
\bibliography{sn-bibliography}